\documentclass{article}
\usepackage[T1]{fontenc} 
\usepackage[utf8]{inputenc} 
\usepackage{ismir,amsmath,cite}
\usepackage[hyphens]{url}
\usepackage{graphicx}
\usepackage{color}

\usepackage{lineno}
\usepackage{enumitem}

\usepackage{amssymb}
\usepackage{booktabs}
\usepackage{multirow}

\title{Toward a More Complete OMR Solution}

\multauthor
{Guang Yang$^1$ \hspace{.5cm} Muru Zhang$^1$ \hspace{.5cm} Lin Qiu$^1$ \hspace{.5cm} Yanming Wan$^1$ \hspace{.5cm} Noah A. Smith$^{1,2}$}{
  $^1$Paul G. Allen School of Computer Science \& Engineering, University of Washington, United States\\
  $^2$Allen Institute for Artificial Intelligence, United States \\
{\tt \{\href{mailto:gyang1@cs.washington.edu}{gyang1},\href{mailto:nasmith@cs.washington.edu}{nasmith}\}@cs.washington.edu}
}

\def\authorname{G. Yang, M. Zhang, L. Qiu, Y. Wan, and N. A. Smith}

\usepackage[bookmarks=false,pdfauthor={\authorname},pdfsubject={\papersubject},hidelinks]{hyperref}

\begin{document}

\maketitle
\begin{abstract}
Optical music recognition (OMR) aims to convert music notation into digital formats. 
One approach to tackle OMR is through a multi-stage pipeline, where the system first detects visual music notation elements in the image (object detection) and then assembles them into a music notation (notation assembly). Most previous work on notation assembly unrealistically assumes perfect object detection. In this study, we focus on the MUSCIMA++ v2.0 dataset, which represents musical notation as a graph with pairwise relationships among detected music objects, and we consider both stages together. First, we introduce a music object detector based on YOLOv8, which improves detection performance. 
Second, we introduce a supervised training pipeline that completes the notation assembly stage based on detection output. We find that this model is able to outperform existing models trained on perfect detection output, showing the benefit of considering the detection and assembly stages in a more holistic way. These findings, together with our novel evaluation metric, are important steps toward a more complete OMR solution.
\end{abstract}
\section{Introduction}\label{sec:introduction}

Optical music recognition (OMR) focuses on converting music notation into digital formats amenable to playback and editing. OMR systems are generally divided into two categories: end-to-end systems (which directly convert the image into music notation) and multi-stage systems. Proposed and refined by \cite{Bainbridge2001, Rebelo2012, Calvo2020}, a standard multi-stage system consists of four stages: preprocessing, music object detection, notation assembly, and encoding. In this study, we focus on the object detection and notation assembly stages.  

MUSCIMA++~\cite{Hajic2017Dataset} suggests representing music notation as a graph where each pair of musical symbols is linked by a binary relationship, allowing for clear notation reconstruction. The authors created a dataset of handwritten scores with a bounding box for each music object and a human-annotated graph of object relationships in each image. 
Notation assembly on MUSCIMA++ can be framed as a set of binary classification decisions to predict the pairwise relationships between music symbols. Most prior research has explored notation assembly with the assumption of perfect detection output~\cite{Pearrubia2023EfficientNA}, but such  assumptions can introduce unwanted biases that deteriorate the performance of the notation assembly system when applied as part of a pipeline. Pacha et al.~\cite{Pacha2019LearningNG} evaluate a notation assembler on realistic detector output, finding some degradation relative to gold-standard objects, but they do not seek to mitigate the problem. 

To improve notation assembly robustness, we propose a training method to complete notation assembly on top of (imperfect) object detection output directly. To have a strong detector to start with, we train YOLOv8~\cite{yolov8_ultralytics} and perform a set of preprocessing steps to adapt the model to the MUSCIMA++ v2.0 dataset. Our detector outperforms previous detectors on MUSCIMA++ v2.0~\cite{Zhang2023ADF} by 2.4\%, 
establishing a solid foundation for notation assembly.

Traditional evaluation methods, which perform notation assembly over all pairs of ground-truth objects and report an F1 score or a precision-recall curve, become inadequate when the input objects come from imperfect detection. We propose an end-to-end evaluation metric, called Match+AUC,
that accounts for both detection errors and assembly errors by first matching detected objects with their ground-truth counterparts before assessing notation assembly accuracy.  It complements metrics that evaluate pipeline components individually. 

Our code for reproducing all of the experiments is publicly available at \url{https://github.com/guang-yng/completeOMR}.
\section{Multi-Stage OMR}\label{sec:methodology}

\begin{figure*}
    \centering
    \includegraphics[width=0.8\textwidth]{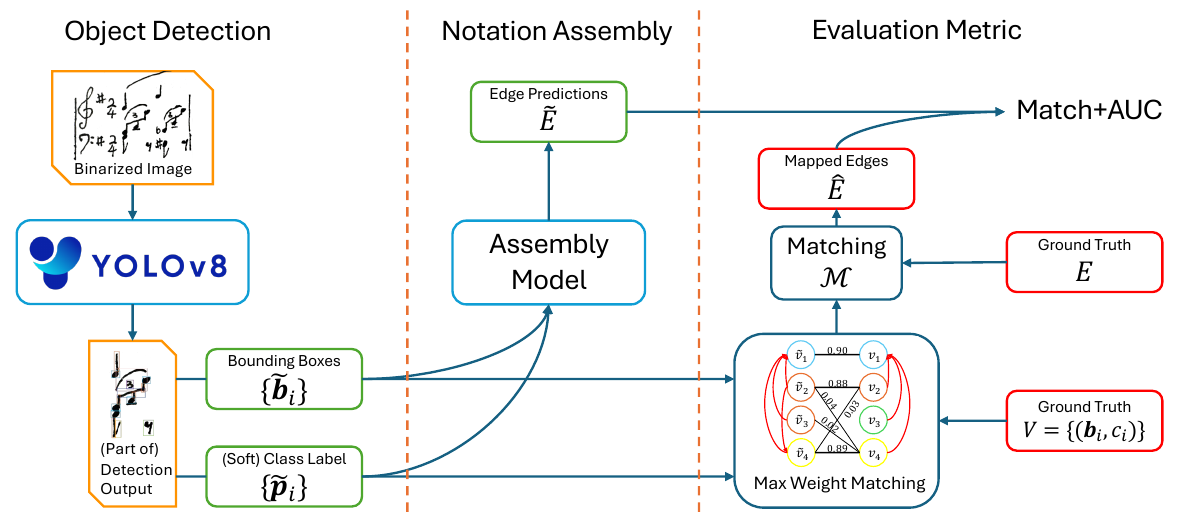}
    \caption{
    An overview of our OMR pipeline, highlighting key components: object detection, notation assembly, and evaluation metric. Detailed explanations of each component can be found in Subsections~\ref{subsec:detection}, \ref{subsec:assembly}, and \ref{subsec:e2e_eval} respectively.
    \label{fig:overview}}
    \vspace{-0.5cm}
\end{figure*}

\begin{figure}
 \includegraphics[width=\columnwidth]{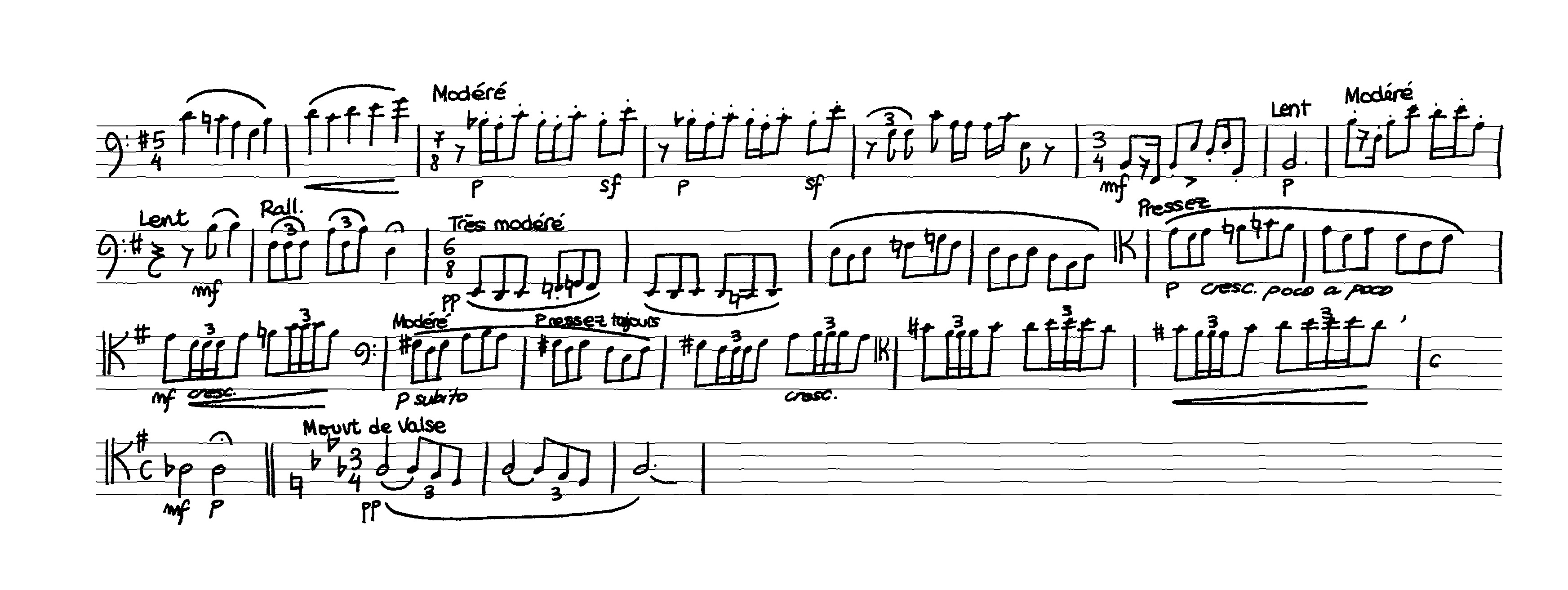}
 \vspace{-0.5cm}
 \caption{Example of a music image (binarized) extracted from the MUSCIMA++ dataset.}
 \label{fig: example data}
 \vspace{-0.5cm}
\end{figure}

We focus on the MUSCIMA++ v2.0 dataset~\cite{Hajic2017Dataset} and follow its multi-stage pipeline for the OMR system. This dataset includes 140 high-resolution annotated images out of 1,000 images from the CVC-MUSCIMA dataset~\cite{Forns2012CVCMUSCIMAAG}. It contains 91,254 symbol-level annotations and 82,247 relationship annotations between symbol pairs by human annotators.  These annotations span 163 distinct classes of music symbols. \figref{fig: example data} shows an example from this dataset.

As the MUSCIMA++ dataset provides symbol-level pairwise relationships, it allows study of two stages of the pipeline: (i) detection and (ii) assembly. In (i), given an image as input, an object detector is used to extract all music symbols in the image, denoted as the set $V = \{v_i\}_i$, where $v_i = (\mathbf{b}_i, c_i)$ is a tuple of a bounding box and a class label. Each pair of music symbols $(v_i, v_j)$ is then fed into (ii) the notation assembly model to predict whether or not there exists a relationship between them. The notation assembly stage can be framed as an edge prediction problem where the model needs to output a set of edges $E$ to get a directed graph $G = (V, E)$.  MUSCIMA++ defines a grammar over all possible music symbol classes so that the direction of an edge is uniquely determined by the class labels $(c_i, c_j)$ of the vertices $(v_i, v_j)$. Consequently, the edge prediction problem can be reduced to predicting an undirected graph.
The authors of \cite{Hajic2017Dataset} argue that such a graph $G$ enables straightforward reconstruction of the full symbolic music notation, so we do not consider the decoding process after (i) and (ii) in this work.

In previous works, the two stages are considered separately, either focusing on object detection, without fully analyzing its effect on downstream notation assembly~\cite{Pacha2018HandwrittenMO, Zhang2023ADF, Pacha2018ABF}; or focusing on notation assembly and assuming perfect detection input during training~\cite{Pacha2019LearningNG, Pearrubia2023EfficientNA}. This raises the question of whether the best object detector is a good fit for the best notation assembly model. To investigate, we developed an end-to-end metric that evaluates the performance of the entire pipeline, as explained in Section~\ref{subsec:e2e_eval}. We found that, compared with our approach where both stages are considered together---specifically, where the notation assembly model is trained using the output of the object detector---treating the two stages separately leads to poorer results.

\section{Methodology}\label{sec:Methodology}
We describe our method for each stage, and how we connect the two stages together and evaluate the entire pipeline.
\figref{fig:overview} shows an overview of our methods.

\subsection{Music Symbol Detection}\label{subsec:detection}
A music object detection system analyzes an image to identify each music object it contains, providing both the bounding box and class label for every detected object~\cite{Pacha2018HandwrittenMO}. Traditionally, this process would begin with an initial stage of image preprocessing, typically aimed at removing staff lines, followed by a second stage focusing on the segmentation and classification of symbols. Thanks to recent advances in computer vision, there are mature solutions for image preprocessing and staff line removal, allowing us to treat it as a largely solved problem~\cite{konwer2018staff, CalvoZaragoza2017StafflineDA, Gallego2017StafflineRW}. 
In our case, MUSCIMA++ provides us with staff line removed images as input, so we directly build our detectors on top of these images.

Following the work of Zhang et al.~\cite{Zhang2023ADF}, we adopt a convolutional neural network-based approach for page-level object detection of handwritten music notes, opting for this approach over segmentation-based methods, because segmentation-based methods often struggle with overlapping symbols. 
We choose YOLOv8~\cite{yolov8_ultralytics}, which is the latest version of YOLO~\cite{yolo}, due to its superior performance on traditional computer vision tasks. Compared to YOLOv4~\cite{yolov4}, which is used by Zhang et al.~\cite{Zhang2023ADF}, YOLOv8 has a new loss function and a new anchor-free detection head, achieving higher performance on various detection tasks.  YOLOv8 has not yet, to our knowledge, been applied to OMR. 
Furthermore, since the images of handwritten music notation in MUSCIMA++ have high resolution and music objects are drastically different from the objects considered in computer vision research, directly applying the training strategy of YOLOv8 doesn't work well. We follow previous works~\cite{Zhang2023ADF, BARO20191, Rebelo2012} to crop images into small snippets during the training stage to alleviate this issue. 
Specifically, we randomly crop the images during training and compactly segment the image during inference.
More details are presented in Section~\ref{sec:filter}.

\begin{figure}[tb]
    \centering
    \includegraphics[width=0.8\columnwidth]{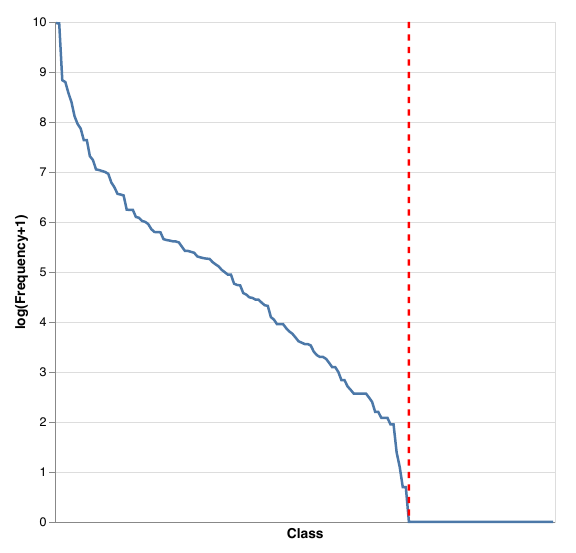}
    \caption{Frequencies of different classes in the dataset, from most- to least-frequent. A long-tailed distribution with 48 classes on the right of the red line that never appear. The $y$-axis shows the value of $\ln{(\text{frequency}+1)}$.  
    The top-5 classes are \texttt{stem}, \texttt{nodeheadFull}, \texttt{ledgerLine}, \texttt{beam}, and \texttt{staffSpace}.
    \label{fig:frequency}}
    \vspace{-0.5cm}
\end{figure}

The MUSCIMA++ v2.0 dataset includes 163 object classes in total, covering a large variety of notation. However, most of the classes scarcely appear and barely affect the replayability of the OMR output (e.g., the construction of a MIDI file encoding the score). The distribution of classes is shown in \figref{fig:frequency}; 48 classes never appear in the entire dataset. Given this, we manually remove these 48 classes along with some other rare classes, leading to a subset of 73 attested ``essential'' classes that are observed in the dataset. 
To get a direct comparison with previous methods, while also keeping a focus on essential classes, we report results using both the full class set and essential classes only. Meanwhile, we also report results on the 20 ``primitive'' classes selected for evaluation by Zhang et al.~\cite{Zhang2023ADF}.

\subsection{Notation Assembly}\label{subsec:assembly}
The notation assembly model takes a pair of nodes as input, and gives a binary output indicating whether there is a relationship between them. An intuitive method is to first concatenate the features of two nodes, and then pass the pair as a single feature vector through a series of layers of a multi-layer perceptron (MLP). A sigmoid function $\sigma$ is applied at the end to output the probability that there exists a relationship.
\begin{equation}
    \hat{e}_{ij}=\sigma(\phi_\text{MLP}([v_i, v_j])) \label{eq:MLP}
\end{equation}

As notation assembly is essentially binary classification, we use binary cross-entropy as our loss function: 
\begin{equation*}
    \mathcal{L}_\text{BCE}(\hat{e}_{ij})=-e_{ij}\log(\hat{e}_{ij}) - (1-e_{ij})\log(1-\hat{e}_{ij}).
\end{equation*}

We adopt the input feature design in~\cite{Pearrubia2023EfficientNA}, where each $v_i$ is represented by its 4-dimensional bounding box coordinates and the class label. The class label is passed to an embedding layer with $x$ dimensions. Therefore, the input to MLP will be a $(4+x) \times 2$ dimensional vector. 

Existing work assumes perfect detection output; therefore, the input bounding box and class label are the ground-truth information. While previous work has attempted to manually perturb the bounding box as a test of robustness, such perturbations don't reflect the kind of errors that might arise in a practical object detector.

To ensure our notation assembly system can adapt to errors introduced in the detection stage, we propose a supervised training pipeline that directly trains the assembly model on detection output $\tilde{V}$. Since most of the time $\tilde{V} \neq V$, we can't directly use the ground truth $E$ as the supervision signal.

To deal with this issue, we construct a maximum weight matching $M$ in the bipartite graph $G_M = (\tilde{V}, V)$ and build $\hat{E}$ for supervising our notation assembly model. We describe the detail of our matching procedure in Section~\ref{subsec:e2e_eval}, where it is also employed in evaluation. We adopt the edges from the ground truth according to our matching. Given a pair $(\tilde{v}_i, v_k)\in M$ and an edge $(v_k, v_h)\in E$, we add $(\tilde{v}_i, \tilde{v}_j)$ to $\hat{E}$ if $(\tilde{v}_j, v_h)\in M$. Our method essentially builds a training set for the detection output that is in the same format as the ground-truth, allowing seamless training and evaluation. 

\subsection{End-to-End Evaluation}\label{subsec:e2e_eval}

 The main challenge of OMR evaluation is finding the edit distance between two music scores under some particular representation (e.g., XML format~\cite{foscarin2019diff}). Haji\v{c}~\cite{hajivc2018case} argued that intrinsic evaluation is needed to decouple research of OMR methods from individual downstream use-cases, since specific notation formats change much faster than music notation itself. Some works have taken steps to analyze the complexity of standard music notation~\cite{byrd2015towards} and propose common music representation formats~\cite{torras2023common}.

As a general system consisting several modules, we seek to  also evaluate our OMR pipeline  holistically, without a specific focus on what the downstream processing will be. We therefore propose a novel matching-based evaluation metric  to assess predictions that include errors from the detection stage.  For the same reason that we had to adapt ground-truth edges to create training data for the notation assembly model~($\tilde{V} \neq V$), we cannot straightforwardly use the ground-truth graph to evaluate notation assembly. Our metric finds a matching between a test instance's predicted objects and those in the ground-truth object detection,  and then uses this as a bridge to evaluate the edges returned by the notation assembly module.

The results reported by Pacha et al.~\cite{Pacha2019LearningNG} are the sole benchmark for assessing a notation assembly model using detected symbols. To address the matching issue between $\tilde{V}$ and $V$, Pacha et al. employ a rule-based method, considering two objects identical if they belong to the same class and their intersection over union is at least 50\%. However, this greedy matching approach is inadequate, as inaccuracies in symbol detection cannot be compensated for by the notation assembly model. Furthermore, Pacha et al. use conventional precision/recall metrics with a hard decision boundary, which fails to capture the overall performance of the model comprehensively. To resolve these issues, we propose a complementary metric based on a global optimal matching and area under the precision-recall curve.

Formally, we denote $\tilde{V} = \{\tilde{v}_1, \tilde{v}_2, \cdots, \tilde{v}_{\tilde{n}}\}$ as the set of symbols obtained from an object detection model, where $\tilde{v}_i = (\tilde{\mathbf{b}}_i, \mathbf{p}_i)$ is a tuple of a bounding box $\tilde{\mathbf{b}}_i \in \mathbb{R}^4$ and a probability distribution vector $\mathbf{p}_i \in \mathbb{R}^{C}$ over all symbol classes.
A notation assembly prediction on $\tilde{V}$ would be an edge set $\tilde{E} = \{\tilde{e}_1, \tilde{e}_2, \cdots, \tilde{e}_{\tilde{m}}\}$ where each edge $\tilde{e}_i$ is a tuple of two vertices.
Similarly, we denote the ground truth notation graph as $G = (V, E)$ with $V = \{v_1, v_2, \cdots, v_n\}, v_i = (\mathbf{b}_i, c_i), E = \{e_1, e_2, \cdots, e_m\}$, where $\mathbf{b}_i \in \mathbb{R}^4$ is a bounding box and $c_i \in \{1, 2, \cdots, C\}$ is a symbol class label.  

We first construct a complete weighted bipartite $(\tilde{V}, V)$ where the weight for edge $(\tilde{v}_i, v_j)$ is $w_{ij} = \mathrm{IoU}(\tilde{\mathbf{b}}_i, \mathbf{b}_j) \cdot \mathbf{p}_{i, c_j}$. Here, $\mathrm{IoU}$ is the intersection-over-union between the area occupied by the two boxes, defined as:
\[\mathrm{IoU}(\mathbf{b}_i, \mathbf{b}_j) = \frac{\mathrm{Area}(\mathbf{b}_i \cap \mathbf{b}_j)}{\mathrm{Area}(\mathbf{b}_i \cup \mathbf{b}_j)}.
\]
Based on this bipartite graph, we find the maximum weighted matching $M$ using the implementation described in \cite{Course} and filter the ``weak" matching edges with weight $w_{ij}$ less than a threshold $T_{\mathrm{match}}$
to get the matching function $\mathcal{M}: V \to \tilde{V} \cup \{\varnothing\}$:
\[
\mathcal{M}(v_j) = \begin{cases}
    \tilde{v}_i, \text{ if } (\tilde{v}_i, v_j) \in M \text{ and } w_{ij} > T_{\mathrm{match}},\\
    \varnothing, \text{ otherwise.}
\end{cases}
\]
Here, $T_{\mathrm{match}}$ is a filtering threshold for matching and 
we set it to $0.05$ without tuning. 

\begin{figure}[t]
    \centering
    \includegraphics[width=0.9\columnwidth]{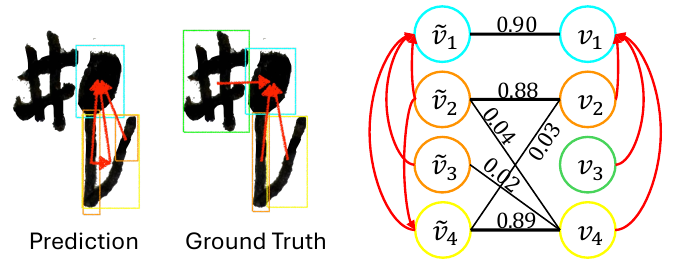}
    \caption{
    An example of detected objects and predicted graph, alongside ground truth.
    At the right is the constructed bipartite graph (zero-weight edges not shown). Thick edges represent the matching function $\mathcal{M}$ induced by the matching algorithm. In our notation, $E = \{(v_2, v_1), (v_3, v_1), (v_4, v_1)\}$ and the matching function maps $v_1$ to $\tilde{v}_1$, $v_2$ to $\tilde{v}_2$ and $v_4$ to $\tilde{v_4}$.
    Therefore, $\hat{E} = \{(\tilde{v}_2, \tilde{v}_1), (\tilde{v}_4, \tilde{v}_1)\}$. Because $\tilde{E} = \{(\tilde{v}_2, \tilde{v}_1), (\tilde{v}_2, \tilde{v}_4), (\tilde{v}_3, \tilde{v}_1), (\tilde{v}_4, \tilde{v}_1)\}$, we get a precision of 0.5 and recall of 1.0.}
    \vspace{-0.5cm}
    \label{fig:metric_example}
\end{figure}

After getting the matching function, the ground truth assembly edges are naturally mapped back to edges between predicted vertices. The mapped edge set $\hat{E} = \{(\mathcal{M}(v_i), \mathcal{M}(v_j)) \mid (v_i, v_j) \in E, \mathcal{M}(v_i) \neq \varnothing, \mathcal{M}(v_j) \neq \varnothing\}$ represents a ground truth edge set on detected vertices, which can be used to evaluate predictions $\tilde{E}$ to get a precision and recall.  An example is shown in \figref{fig:metric_example}.

Most notation assembly models predict a probability of the existence of an edge $(v_i, v_j)$, and the probability is further compared with a threshold $T_{\mathrm{predict}}$ to determine whether $(v_i, v_j)$ belongs to the prediction set $\tilde{E}$. 
By adjusting the model prediction threshold $T_{\mathrm{predict}}$,
we can get a series of predictions $\{\tilde{E}_1, \tilde{E}_2, \cdots \}$ and therefore derive a series of precision-recall pairs, which are used to estimate the area-under-the-curve (AUC) score.  
We refer to the full evaluation metric as ``Match+AUC.''

``Match+AUC'' is an end-to-end evaluation metric for the OMR pipeline with following advantages:
\begin{itemize}[leftmargin=*]
    \item ``Match+AUC'' accounts for model performance in both the object detection and notation assembly stages. To be specific, given an object detector's output, a notation assembly model will achieve a higher score if it predicts no edges among redundant objects, since connecting redundant nodes into the assembly graph would greatly affect the final output music score.  Also, for the same assembly model, a worse object detector would generate a large amount of redundant and inaccurate objects, making it very hard for the assembly model to distinguish them. 
    \item Instead of a hard rule-based matching used in past methods, ``Match+AUC'' creates a comprehensive matching among detected symbols and ground truth symbols, making the final score more accurate and sensitive.
    \item ``Match+AUC'' evaluates the model using the area under the precision-recall curve, which summarizes performance across a range of threshold choices that could be made by a downstream module or a system user.
\end{itemize} 
We believe that our novel ``Match+AUC'' is a compelling tool for analyzing OMR pipelines that is complementary to existing approaches. 

\section{Implementation Details}\label{sec:experiment}


\subsection{Music Symbol Detection}

\subsubsection{Model Details}

We finetune the ``large'' version of YOLOv8 (YOLOv8l), an object detection model pre-trained on the COCO dataset~\cite{COCO}, on MUSCIMA++ v2.0 for music object detection.
The model consists of 43.7M parameters and is capable of detecting object bounding boxes and generating corresponding class distributions. The input image size of our model is set to 640.

\subsubsection{Training}
\label{sec:filter}
We used the MUSCIMA++ v2.0 dataset to train and evaluate the music symbol detection model~\cite{Hajic2017Dataset}. The images are binarized (pixels are 0/1-valued) and in a size of approximately 3500 $\times$ 2000 pixels. For simplicity, we use images with staff lines removed. 
Additionally, following the exact method described in \cite{Pacha2019LearningNG}, we split the dataset into 60\% training data, 20\% validation data, and 20\% test data.
To effectively train YOLOv8 on these dense images involving many small annotations, which include augmentation dots and piano pedal markings, we have to reduce the image size. Therefore, following the methods used by~\cite{Zhang2023ADF}, given a large music score image, we randomly sample 14 
crops of size $1216 \times 1216$ and then resize them to $640 \times 640$ to fit the YOLOv8 input requirement. 

We fine-tune the YOLOv8 model for 500 epochs with a batch size of 8. We use the AdamW optimizer~\cite{loshchilov2018decoupled} with a learning rate of $5.5\times 10^{-5}$ and a momentum of $0.9$, which are automatically set by the YOLOv8 codebase~\cite{yolov8_ultralytics}.
During training, we use the early stopping strategy with a patience of 100 epochs. We keep the checkpoint with the highest validation performance as our final model.

\subsubsection{Inference}
\begin{figure}[tb]
\centering
 \includegraphics[width=0.8\columnwidth]{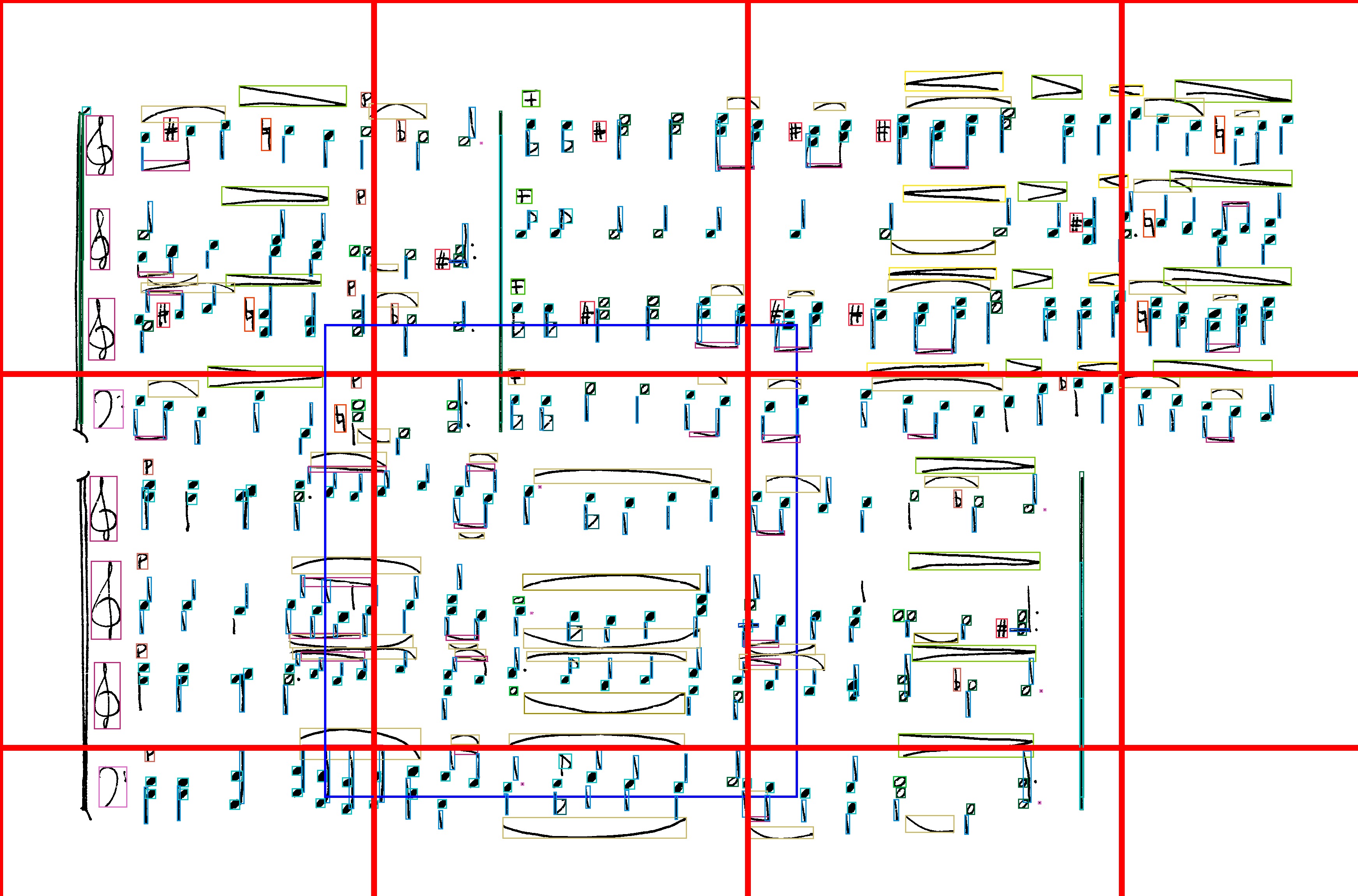}
 \caption{Example of music symbol detection segments for inference. The thick red line indicates the primary cropped area, while the thick blue line represents an extended cropped section designed to include partial symbols that may extend beyond the main cropped area. For better visualization, we only show the extended area of one image crop. Image crops on the right and bottom border of the page are padded to fit into YOLOv8.}
 \label{fig: example crop}
 \vspace{-0.5cm}
\end{figure}

Since our detector is trained on cropped data, during the inference stage, we also need to segment the large images into smaller segments. 
However, partial objects at the edges of these crops would be hard to detect since the model can't see the full object. 
To resolve this issue, we extend every crop with a margin, which serves as a context for each image.
The cropping is visualized in an example in \figref{fig: example crop}. 
We then perform symbol detection on each extended crop and consolidate the detection results.
To make sure the objects on the edges are only detected once,
overlapping bounding boxes are filtered based on their Intersection over Union (IoU) overlap rate. 

\subsection{Notation Assembly}\label{subsec:assembly detail}

\subsubsection{Model Details}
We use a 4-layer MLP for $\phi_\text{MLP}$, where the two hidden layers both have a hidden dimension of 32. The embedding dimension for the symbol class is also set to be 32. We use ReLU~\cite{agarap2019deep} as the activation function. 

\subsubsection{Training}
Again we use the MUSCIMA++ v2.0 dataset to train and evaluate the notation assembly model~\cite{Hajic2017Dataset}.
Following previous work~\cite{Pacha2019LearningNG, Pearrubia2023EfficientNA}, we balance the positive and negative pairs in the training set by filtering out the pairs of nodes that are too distant from each other since they are unlikely to be connected. Before feeding the bounding box coordinates to the model, 
we normalize them by the image width while keeping the aspect ratio fixed, so that all of the $x$-coordinate values fit in the range of $[-1, 1]$. 

We train our models for 200 epochs with batch size 256, and use Adam optimizer~\cite{kingma2017adammethodstochasticoptimization} with a learning rate of 0.0001. We evaluate our model every 20 epochs and pick the checkpoint with highest validation Match+AUC as our final model. All of the experiments are conducted with five different random seeds.

In our experiments, we consider three methods for training the notation assembly model:
\begin{itemize}[leftmargin=*]
    \item A \textbf{baseline}, which uses the ground-truth object lists provided in the MUSCIMA++ dataset to train the notation assembly model. This is the setup used in~\cite{Pearrubia2023EfficientNA}.
    \item A \textbf{pipeline}, which runs the music object detection model on the images to construct the training set for the notation assembly model, as discussed in Section~\ref{subsec:assembly}. 
    \item A ``soft'' variant of the pipeline, where we replace the embedding layer for the symbol class with a linear layer that maps the symbol class probabilities outputted from the music object detection model to a 32-dimensional vector. Note that this linear layer will have the same parameter count (number of classes multiplied by the hidden dimension) as the replaced embedding layer.
\end{itemize}

\subsubsection{Inference}
Since we consider both stages together, the input to the notation assembly stage should correspond to the output of the object detection stage. As described in Section~\ref{subsec:assembly}, the detection output is converted into $(V', E')$. We then pass each pair of nodes to the notation assembly model, and feed the result into our evaluation function. We hypothesize that this realistic setup introduces a distribution shift to the model that was trained on the ground-truth objects and we will make the comparison in Section~\ref{sec:result}.

\section{Experiments}\label{sec:result}

In this section, we first report the performance of our music symbol detection model. Then, we compare the performance of different notation assembly training pipelines using the evaluation metric described in Section~\ref{subsec:e2e_eval}.

\begin{table*}[h]
\small
\centering
\begin{tabular}{lrlrr}
\toprule
Models & \# Classes &&  mAP (\%) & Weighted mAP (\%) \\ 
\midrule
YOLOv8 + cropping (ours) & 163 &(all)  & 84.79 & 92.67\hphantom{$^\dagger$}\\ 
YOLOv8 + cropping (ours) & 73 & (essential) & 85.67 & 89.96\hphantom{$^\dagger$}\\ 
\midrule
YOLOv8 + cropping (ours) & 20&  & \textbf{94.22} &  \textbf{95.72}\hphantom{$^\dagger$} \\ 
YOLOv4 + CBAM \cite{Zhang2023ADF} & 20& & 91.8\hphantom{0} & 94.56$^\dagger$ \\ 
PP-YOLO-V2 \cite{Zhang2023ADF} & 20 && 91.1\hphantom{0} & --\hphantom{$^\dagger$} \\
YOLO-X \cite{Zhang2023ADF} & 20& & 90.4\hphantom{0} & --\hphantom{$^\dagger$} \\ 
YOLOv4 \cite{Zhang2023ADF} & 20 && 89.1\hphantom{0} & --\hphantom{$^\dagger$} \\
Faster R-CNN \cite{Zhang2023ADF}& 20 && 86.2\hphantom{0} & --\hphantom{$^\dagger$} \\
\bottomrule
\end{tabular}
\caption{Object detection results on test set. ``mAP'' is mean average precision. We compared it with results reported by~\cite{Zhang2023ADF}. The lower block is included for comparability with the 20-class setting from past work. $\dagger$: Value computed from average precision per class reported in \cite{Zhang2023ADF}.
} 
\label{tab: detection result}
\end{table*}

\begin{table*}
\small
    \centering
    \begin{tabular}{lc|c}
    \toprule
    \multirow{2}{*}{Models} & \multirow{2}{*}{\# Classes} & Match+AUC \\
     & & Average\qquad S.D. \\
    \midrule
    MLP baseline (train on ground truth objects) & 73 &  $92.44\pm 0.24$ \\
    \quad + pipelined training (ours) & 73 & $93.09\pm0.16$ \\
    \quad + pipelined training + soft label (ours) & 73 & $\textbf{95.00}\pm0.18$ \\
    \midrule
    MLP baseline (train on ground truth objects) &  163 &  $83.97\pm 3.04$ \\
    \quad + pipelined training (ours) & 163 & $85.76\pm 0.42$ \\
    \quad + pipelined training + soft label (ours) & 163 & $\textbf{87.10}\pm1.19$ \\
    \bottomrule
    \end{tabular}
    \caption{Multi-stage system results (test set) using our Match+AUC metric. \label{tab:full-results}}
\end{table*}

\subsection{Music Symbol Detection}\label{subsec:detection result}
Following the evaluation protocols of the Pascal VOC challenge~\cite{Everingham2015}, which are used by previous methods~\cite{Pacha2018HandwrittenMO, Pacha2018ABF, Zhang2023ADF}, we present both the mean average precision (mAP) and the weighted mean average precision,
as detailed in \tabref{tab: detection result}. 
To elaborate, a predicted bounding box $\tilde{\mathbf{b}}_i$ is thought to be a true positive only if $\mathrm{IoU}(\tilde{\mathbf{b}}_i, \mathbf{b}_j) > 0.5$ for some ground truth box $\mathbf{b}_j$.
Then, average precision (AP) computes the area under the precision-recall curve, providing a single value that encapsulates the model's precision and recall performance.
The weighted/unweighted mean Average Precision (mAP) extends the concept of AP by calculating the average AP values across multiple object classes, taking into account the number of occurrences of each class in a weighted or unweighted manner.
Our experiments are conducted with the MUSCIMA++ v2.0 dataset, while the authors of most previous methods~\cite{Pacha2018HandwrittenMO, Pacha2018ABF} have only tested their models on MUSCIMA++ v1.0. This introduces a misalignment between our results.
Thanks to Zhang et al.~\cite{Zhang2023ADF}, who provided reproduced results of most previous methods on MUSCIMA++ v2.0, we directly report their reproduced results in the table.

Our model outperforms Zhang et al.'s method on their selected 20 classes by 2.4\% (mAP, absolute), likely due to the improvements in YOLOv8 compared to v4. 

\subsection{Notation Assembly}

In this section, we complete the multi-stage OMR system by chaining different notation assembly models to the best music object detection model we trained in Section~\ref{subsec:detection result}. We use the metric we designed in Section~\ref{subsec:e2e_eval} to report the end-to-end performance of the OMR system. 

In \tabref{tab:full-results}, we compare the notation assembly systems trained with baseline training, pipelined training, and soft pipelined training as described in Section~\ref{subsec:assembly detail}.
We found that pipelined training improves the Match+AUC score by  0.65\% (essential) and 1.79\% (all), absolute, and incorporating the soft class label further increases the performance by 1.91\% (essential) and 1.34\% (all), absolute.
Training the notation assembly model on the detection model output and using the soft label probability to represent the class information, we are able to improve the Match+AUC of the OMR system by 3.13\%.
We hypothesize that pipelined training helps the assembly model adapt to any inaccuracies our object detector has, and incorporating the soft class labels enables the assembly model to consider alternative class labels, not just those chosen by the object detector.

\section{Conclusion and Future Work}\label{sec:conclusion}
In our study, we reconsider a multi-stage OMR pipeline built and evaluated using the MUSCIMA++ dataset. We first propose a state-of-the-art music symbol detector, serving as a strong preprocessor for the notation assembly stage. We then propose a training pipeline in which notation assembly is learned from imperfect object detection outputs (rather than ground-truth objects), which leads to higher performance. Finally, we introduce an evaluation score, Match+AUC, which can jointly consider the error in both detection and assembly stages, allowing evaluation of the two stages together.

Match+AUC is not restricted to being an evaluation metric. Future research could explore the application of Match+AUC within a joint training objective function for both the object detection and notation assembly stages. This approach would enable the entire model to be optimized for retrieving a globally optimal music notation graph. 

In this study, we focus on the object detection and notation assembly stages in the OMR pipeline. Progress on the encoding stage is also required for a complete OMR solution; while the music notation graph arguably contains the essential information for recovering a score \cite{Hajic2017Dataset}, conversion of such graphs into standard formats remains unsolved.  
Exploring methods to effectively convert music notation graphs into standard formats could be a valuable research direction to achieve a more complete OMR solution.

\section{Acknowledgments} 
The authors wish to express our deepest gratitude to all creators of the public OMR datasets for their dedication and generosity in collecting and sharing these invaluable resources. We extend our sincere thanks to Carlos Peñarrubia for his assistance in clarifying questions regarding the reproduction of their method. We are also grateful to Tim Althoff for his insightful comments on our evaluation metric. Special thanks go to Victoria Ebert and Teerapat Jenrungrot for providing us with essential materials in the OMR field. Finally, we sincerely appreciate the constructive reviews, which have significantly enhanced the rigor and completeness of this paper.

\bibliography{ISMIRtemplate}

\begin{thebibliography}{10}
\providecommand{\url}[1]{#1}
\csname url@samestyle\endcsname
\providecommand{\newblock}{\relax}
\providecommand{\bibinfo}[2]{#2}
\providecommand{\BIBentrySTDinterwordspacing}{\spaceskip=0pt\relax}
\providecommand{\BIBentryALTinterwordstretchfactor}{4}
\providecommand{\BIBentryALTinterwordspacing}{\spaceskip=\fontdimen2\font plus
\BIBentryALTinterwordstretchfactor\fontdimen3\font minus \fontdimen4\font\relax}
\providecommand{\BIBforeignlanguage}[2]{{%
\expandafter\ifx\csname l@#1\endcsname\relax
\typeout{** WARNING: IEEEtran.bst: No hyphenation pattern has been}%
\typeout{** loaded for the language `#1'. Using the pattern for}%
\typeout{** the default language instead.}%
\else
\language=\csname l@#1\endcsname
\fi
#2}}
\providecommand{\BIBdecl}{\relax}
\BIBdecl

\bibitem{Bainbridge2001}
\BIBentryALTinterwordspacing
D.~Bainbridge and T.~Bell, ``The challenge of optical music recognition,'' \emph{Computers and the Humanities}, vol.~35, pp. 95--121, 05 2001. [Online]. Available: \url{https://doi.org/10.1023/A:1002485918032}
\BIBentrySTDinterwordspacing

\bibitem{Rebelo2012}
\BIBentryALTinterwordspacing
A.~Rebelo, I.~Fujinaga, F.~Paszkiewicz, A.~R.~S. Marcal, C.~Guedes, and J.~S. Cardoso, ``Optical music recognition: state-of-the-art and open issues,'' \emph{International Journal of Multimedia Information Retrieval}, vol.~1, no.~3, pp. 173--190, Oct. 2012. [Online]. Available: \url{https://doi.org/10.1007/s13735-012-0004-6}
\BIBentrySTDinterwordspacing

\bibitem{Calvo2020}
\BIBentryALTinterwordspacing
J.~Calvo-Zaragoza, J.~{Hajič Jr.}, and A.~Pacha, ``Understanding optical music recognition,'' \emph{ACM Comput. Surv.}, vol.~53, no.~4, jul 2020. [Online]. Available: \url{https://doi.org/10.1145/3397499}
\BIBentrySTDinterwordspacing

\bibitem{Hajic2017Dataset}
\BIBentryALTinterwordspacing
J.~Hajič and P.~Pecina, ``The {MUSCIMA++} dataset for handwritten optical music recognition,'' in \emph{2017 14th IAPR International Conference on Document Analysis and Recognition (ICDAR)}, vol.~01, 2017, pp. 39--46. [Online]. Available: \url{https://doi.org/10.1109/ICDAR.2017.16}
\BIBentrySTDinterwordspacing

\bibitem{Pearrubia2023EfficientNA}
\BIBentryALTinterwordspacing
C.~Peñarrubia, C.~Garrido-Munoz, J.~J. Valero-Mas, and J.~Calvo-Zaragoza, ``Efficient notation assembly in optical music recognition,'' in \emph{{Proceedings of the 24th International Society for Music Information Retrieval Conference}}.\hskip 1em plus 0.5em minus 0.4em\relax ISMIR, Dec. 2023, pp. 182--189. [Online]. Available: \url{https://doi.org/10.5281/zenodo.10265253}
\BIBentrySTDinterwordspacing

\bibitem{Pacha2019LearningNG}
\BIBentryALTinterwordspacing
A.~Pacha, J.~Calvo-Zaragoza, and J.~{Hajič Jr.}, ``Learning notation graph construction for full- pipeline optical music recognition,'' in \emph{Proceedings of the 20th International Society for Music Information Retrieval Conference}.\hskip 1em plus 0.5em minus 0.4em\relax ISMIR, Nov. 2019, pp. 75--82. [Online]. Available: \url{https://doi.org/10.5281/zenodo.3527744}
\BIBentrySTDinterwordspacing

\bibitem{yolov8_ultralytics}
\BIBentryALTinterwordspacing
G.~Jocher, A.~Chaurasia, and J.~Qiu, ``Ultralytics {YOLO}v8,'' 2023. [Online]. Available: \url{https://github.com/ultralytics/ultralytics}
\BIBentrySTDinterwordspacing

\bibitem{Zhang2023ADF}
\BIBentryALTinterwordspacing
Y.~Zhang, Z.~Huang, Y.~Zhang, and K.~Ren, ``A detector for page-level handwritten music object recognition based on deep learning,'' \emph{Neural Computing and Applications}, vol.~35, no.~13, pp. 9773--9787, May 2023. [Online]. Available: \url{https://doi.org/10.1007/s00521-023-08216-6}
\BIBentrySTDinterwordspacing

\bibitem{Forns2012CVCMUSCIMAAG}
\BIBentryALTinterwordspacing
A.~Forn{\'e}s, A.~Dutta, A.~Gordo, and J.~Llad{\'o}s, ``{CVC-MUSCIMA}: a ground truth of handwritten music score images for writer identification and staff removal,'' \emph{International Journal on Document Analysis and Recognition (IJDAR)}, vol.~15, no.~3, pp. 243--251, Sep. 2012. [Online]. Available: \url{https://doi.org/10.1007/s10032-011-0168-2}
\BIBentrySTDinterwordspacing

\bibitem{Pacha2018HandwrittenMO}
\BIBentryALTinterwordspacing
A.~Pacha, K.-Y. Choi, B.~Coüasnon, Y.~Ricquebourg, R.~Zanibbi, and H.~Eidenberger, ``Handwritten music object detection: Open issues and baseline results,'' in \emph{2018 13th IAPR International Workshop on Document Analysis Systems (DAS)}, 2018, pp. 163--168. [Online]. Available: \url{https://doi.org/10.1109/DAS.2018.51}
\BIBentrySTDinterwordspacing

\bibitem{Pacha2018ABF}
\BIBentryALTinterwordspacing
A.~Pacha, J.~{Hajič Jr.}, and J.~Calvo-Zaragoza, ``A baseline for general music object detection with deep learning,'' \emph{Applied Sciences}, vol.~8, 2018. [Online]. Available: \url{https://doi.org/10.3390/app8091488}
\BIBentrySTDinterwordspacing

\bibitem{konwer2018staff}
\BIBentryALTinterwordspacing
A.~Konwer, A.~K. Bhunia, A.~Bhowmick, A.~K. Bhunia, P.~Banerjee, P.~P. Roy, and U.~Pal, ``Staff line removal using generative adversarial networks,'' in \emph{2018 24th International Conference on Pattern Recognition (ICPR)}, 2018, pp. 1103--1108. [Online]. Available: \url{https://doi.org/10.1109/ICPR.2018.8546105}
\BIBentrySTDinterwordspacing

\bibitem{CalvoZaragoza2017StafflineDA}
\BIBentryALTinterwordspacing
J.~Calvo-Zaragoza, A.~Pertusa, and J.~Oncina, ``Staff-line detection and removal using a convolutional neural network,'' \emph{Machine Vision and Applications}, vol.~28, no.~5, pp. 665--674, Aug. 2017. [Online]. Available: \url{https://doi.org/10.1007/s00138-017-0844-4}
\BIBentrySTDinterwordspacing

\bibitem{Gallego2017StafflineRW}
\BIBentryALTinterwordspacing
A.-J. Gallego and J.~Calvo-Zaragoza, ``Staff-line removal with selectional auto-encoders,'' \emph{Expert Systems with Applications}, vol.~89, pp. 138--148, 2017. [Online]. Available: \url{https://doi.org/10.1016/j.eswa.2017.07.002}
\BIBentrySTDinterwordspacing

\bibitem{yolo}
\BIBentryALTinterwordspacing
J.~Redmon, S.~Divvala, R.~Girshick, and A.~Farhadi, ``You only look once: Unified, real-time object detection,'' in \emph{2016 IEEE Conference on Computer Vision and Pattern Recognition (CVPR)}, 2016, pp. 779--788. [Online]. Available: \url{https://doi.org/10.1109/CVPR.2016.91}
\BIBentrySTDinterwordspacing

\bibitem{yolov4}
\BIBentryALTinterwordspacing
A.~Bochkovskiy, C.-Y. Wang, and H.-Y.~M. Liao, ``{YOLO}v4: Optimal speed and accuracy of object detection,'' 2020. [Online]. Available: \url{https://doi.org/10.48550/arXiv.2004.10934}
\BIBentrySTDinterwordspacing

\bibitem{BARO20191}
\BIBentryALTinterwordspacing
A.~Baró, P.~Riba, J.~Calvo-Zaragoza, and A.~Fornés, ``From optical music recognition to handwritten music recognition: A baseline,'' \emph{Pattern Recognition Letters}, vol. 123, pp. 1--8, 2019. [Online]. Available: \url{https://doi.org/10.1016/j.patrec.2019.02.029}
\BIBentrySTDinterwordspacing

\bibitem{foscarin2019diff}
\BIBentryALTinterwordspacing
F.~Foscarin, F.~Jacquemard, and R.~Fournier-S’niehotta, ``A diff procedure for music score files,'' in \emph{Proceedings of the 6th International Conference on Digital Libraries for Musicology}, ser. DLfM '19, 2019, p. 58–64. [Online]. Available: \url{https://doi.org/10.1145/3358664.3358671}
\BIBentrySTDinterwordspacing

\bibitem{hajivc2018case}
J.~{Hajič Jr.}, ``A case for intrinsic evaluation of optical music recognition,'' \emph{International Workshop on Reading Music Systems}, 2018.

\bibitem{byrd2015towards}
\BIBentryALTinterwordspacing
D.~Byrd and J.~Simonsen, ``Towards a standard testbed for optical music recognition: Definitions, metrics, and page images,'' \emph{Journal of New Music Research}, vol.~44, 07 2015. [Online]. Available: \url{https://doi.org/10.1080/09298215.2015.1045424}
\BIBentrySTDinterwordspacing

\bibitem{torras2023common}
P.~Torras, S.~Biswas, and A.~Forn{\'e}s, ``The common optical music recognition evaluation framework,'' \emph{arXiv preprint arXiv:2312.12908}, 2023.

\bibitem{Course}
\BIBentryALTinterwordspacing
D.~F. Crouse, ``On implementing {2D} rectangular assignment algorithms,'' \emph{IEEE Transactions on Aerospace and Electronic Systems}, vol.~52, no.~4, pp. 1679--1696, 2016. [Online]. Available: \url{https://doi.org/10.1109/TAES.2016.140952}
\BIBentrySTDinterwordspacing

\bibitem{COCO}
\BIBentryALTinterwordspacing
T.-Y. Lin, M.~Maire, S.~Belongie, J.~Hays, P.~Perona, D.~Ramanan, P.~Doll{\'a}r, and C.~L. Zitnick, ``Microsoft {COCO}: Common objects in context,'' in \emph{Computer Vision -- ECCV 2014}, D.~Fleet, T.~Pajdla, B.~Schiele, and T.~Tuytelaars, Eds.\hskip 1em plus 0.5em minus 0.4em\relax Cham: Springer International Publishing, 2014, pp. 740--755. [Online]. Available: \url{https://doi.org/10.1007/978-3-319-10602-1_48}
\BIBentrySTDinterwordspacing

\bibitem{loshchilov2018decoupled}
\BIBentryALTinterwordspacing
I.~Loshchilov and F.~Hutter, ``Decoupled weight decay regularization,'' in \emph{International Conference on Learning Representations}, 2019. [Online]. Available: \url{https://openreview.net/forum?id=Bkg6RiCqY7}
\BIBentrySTDinterwordspacing

\bibitem{agarap2019deep}
\BIBentryALTinterwordspacing
A.~F. Agarap, ``Deep learning using rectified linear units ({R}e{LU}),'' 2019. [Online]. Available: \url{https://doi.org/10.48550/arXiv.1803.08375}
\BIBentrySTDinterwordspacing

\bibitem{kingma2017adammethodstochasticoptimization}
\BIBentryALTinterwordspacing
D.~P. Kingma and J.~Ba, ``Adam: A method for stochastic optimization,'' 2017. [Online]. Available: \url{https://arxiv.org/abs/1412.6980}
\BIBentrySTDinterwordspacing

\bibitem{Everingham2015}
\BIBentryALTinterwordspacing
M.~Everingham, S.~M.~A. Eslami, L.~Van~Gool, C.~K.~I. Williams, J.~Winn, and A.~Zisserman, ``The pascal visual object classes challenge: A retrospective,'' \emph{International Journal of Computer Vision}, vol. 111, no.~1, pp. 98--136, Jan. 2015. [Online]. Available: \url{https://doi.org/10.1007/s11263-014-0733-5}
\BIBentrySTDinterwordspacing

\end{thebibliography}

%
%
%
%
%

\end{document}